\documentclass[runningheads]{llncs}
\usepackage{nccmath}

\usepackage{amsmath}

\usepackage{comment}

\usepackage{graphicx}
\begin{document}
\title{Uncertainty categories in medical image segmentation: a study of source-related diversity}
\titlerunning{Uncertainty categories in medical image segmentation}
%
%
\author{Luke Whitbread\inst{1}\orcidID{0000-0001-9960-5592} \and
\\ Mark Jenkinson\inst{1,2}\orcidID{0000-0001-6043-0166}}
\authorrunning{L. Whitbread, M. Jenkinson}
%
\institute{School of Computer Science, The University of Adelaide, Australia \and
Wellcome Centre for Integrative Neuroimaging, University of Oxford, United Kingdom}

\maketitle

\begin{abstract}
Measuring uncertainties in the output of a deep learning method is useful in several ways, such as in assisting with interpretation of the outputs, helping build confidence with end users, and for improving the training and performance of the networks. Several different methods have been proposed to estimate uncertainties, including those from epistemic (relating to the model used) and aleatoric (relating to the data) sources using test-time dropout and augmentation, respectively. Not only are these uncertainty sources different, but they are governed by parameter settings (e.g., dropout rate or type and level of augmentation) that establish even more distinct uncertainty categories. This work investigates how different the uncertainties are from these categories, for magnitude and spatial pattern, to empirically address the question of whether they provide usefully distinct information that should be captured whenever uncertainties are used. We take the well characterised BraTS challenge dataset to demonstrate that there are substantial differences in both magnitude and spatial pattern of uncertainties from the different categories, and discuss the implications of these in various use cases.

\end{abstract}

\section{Introduction}

Anatomical segmentations of magnetic resonance imaging (\textbf{MRI}) scans are commonly used to support a number of clinical and research tasks using either manual or automated methods. While optimising overall measures of success (e.g., Dice/F1 scores) is important for any segmentation task, the uncertainties associated with image segmentations have become a salient field of enquiry for researchers; to (i) improve the quality and interpretability of structural delineations, and (ii) improve trust when applying automated techniques to clinical practice~\cite{ref_dinsdale}. Whilst it is not uncommon in the literature to implicitly consider a single uncertainty distribution, different sources for uncertainties exist (e.g., epistemic and aleatoric). Therefore, a thorough treatment of uncertainties arising from different sources is needed to understand their usefulness, distinctiveness and stability. 

\section{Related Work}
\label{sec:related_work}

Researchers have investigated uncertainties for some time but, more recently, specific epistemic and aleatoric sources of uncertainty have been addressed for deep convolutional networks. Here, two complementary estimation paradigms have emerged; namely, test-time dropout (\textbf{TTD}) for model or epistemic uncertainty and test-time augmentation (\textbf{TTA}) for data or aleatoric uncertainty. The proposed use of Monte Carlo dropout at test-time by Gal and Ghahramani~\cite{ref_gal} overcomes the computational problems of traditional Bayesian approaches, although other approximate models of uncertainty, including Monte Carlo batch normalisation~\cite{ref_taye} and Markov Chain Monte Carlo methods~\cite{ref_neal} have also been proposed. The use of TTD, however, has become more widespread amongst researchers than these other methods. To measure aleatoric uncertainties, Ayhan and Berens proposed the use of TTA~\cite{ref_ayhan}, with Wang and colleagues proposing a systematic approach to assess aleatoric uncertainties in medical imaging~\cite{ref_wang}.  Building upon these paradigms, much research attention has now turned to the incorporation of uncertainties to improve model performance~\cite{ref_ozdemir,ref_herzog,ref_wang2,ref_arega}. 

As the use of uncertainties in imaging pipelines is becoming increasingly common, it is important to perform explorations around which sources of uncertainty are the most beneficial, the most stable and reliable, and the most interpretable. A key question related to this is whether a single uncertainty source and distribution is sufficient for a general use case, or whether multiple sources and distributions should be used.  
Jungo and colleagues~\cite{jungo_1,jungo_2} have addressed issues around reliability and model confidence of epistemic uncertainties for some TTD model architecture strategies and learned aleatoric uncertainty ($\sigma^2$); where $\sigma^2$ is determined using a method proposed by Kendall and Gal~\cite{kendall}. While these and other works (e.g.,~\cite{mehrtash,rousseau}) address uncertainty reliability and model calibration generally, they do not consider different levels of TTD parameters or how aleatoric uncertainties from different underlying sources manifest in observations. Additionally, further important work on modelling uncertainty related to annotator variation has emerged. This utilises latent space encoding techniques derived from the Variational Autoencoder framework~\cite{ref_kingma} to directly model segmentation distributions using a Bayesian approach~\cite{ref_kohl,ref_baumgartner}. These methods provide a mathematically rigorous model for capturing complex voxel dependencies, and though they are currently designed to produce one distribution, they could be modified to capture a number of distinct uncertainty sources, in order to report the most appropriate information for each individual use case.

\section{Methods}

We assess the range and stability of uncertainty measures for epistemic and aleatoric categories by considering various TTD and TTA parameter settings. The parameter space for assessing epistemic uncertainty with TTD is typically less complex than that for aleatoric categories using TTA. While there are different approaches for how dropout is applied structurally to a model, once this is determined, the key variables are the probabilities of the applied dropout. In contrast, TTA has both a number of augmentation types that can be applied (some particularly relevant to MRI), and a number of parameter settings for each type of augmentation. In addition, as is done in model calibration analyses (e.g.,~\cite{mehrtash,rousseau}), we have estimated a prediction error likelihood for the underlying segmentation model, to test how similar it is to other uncertainty types.

\subsection{Epistemic Uncertainties with TTD}
\label{subsec:TTD}

We used the common approach of dropping network filters with a constant global probability. Although researchers sometimes use higher probability parameters at the encoder-decoder junction of a U-net based network, we took the approach of a single probability setting, which still tends to drop more filters at the encoder-decoder junction, given the increasing number of filters with depth.

We have selected 6 TTD probabilities to evaluate the range and stability of epistemic uncertainties: 0.03, 0.06, 0.09, 0.12, 0.15, and 0.40. The first 5 cases represent a range of typical network-wide dropout settings; while the final setting is intended to test what happens when this is set substantially higher.

\subsection{Aleatoric Uncertainties using TTA}
\label{subsec:TTA}

To evaluate the stability of aleatoric uncertainties, we have used 8 TTA cases across three categories common for MRI data: affine transformations (reflecting varying subject orientation), image ghosting and bias-field transforms (both common MRI artefacts). For each category, low and high range parameters have been selected, where the final two cases are transforms composed of all three low or all three high cases. The low settings represent cases that are very likely in practice, whilst the high range settings represent uncommon but plausible cases. These settings are:

\noindent
\textbf{Affine transformation} cases:\\
\hspace*{0.5cm} -- Scaling range: Low$\sim\mathcal{U}(0.98,1.02)$; High$\sim\mathcal{U}(0.80,1.20)$.\\
\hspace*{0.5cm} -- Rotation (degrees): Low$\sim\mathcal{U}(-5,5)$; High$\sim\mathcal{U}(-45, 45)$.\\
\hspace*{0.5cm} -- Image translation (mm): $\sim\mathcal{U}(-5, 5)$ for both cases.

\noindent
\textbf{Ghosting} artefacts (usually caused by subject motion in MRI):\\
\hspace*{0.5cm} -- Intensity strength (max of k-space): Low$\sim\mathcal{U}(0.0,0.15)$; High$\sim\mathcal{U}(0.25,0.75)$.\\
\hspace*{0.5cm} -- Number of ghosts: $\sim\mathcal{U}\{2, 6\}$ for both cases, applied on the 2\textsuperscript{nd} image axis.

\noindent
\textbf{Bias-field} artefacts (smooth intensity changes caused by RF inhomogeneities):\\
\hspace*{0.5cm} -- Maximum polynomial coefficients: Low = 0.2; High = 0.8.\\
\hspace*{0.5cm} -- Polynomial order: 3 for both cases.

\subsection{Dataset}
This work utilises the well characterised Multimodal Brain Tumor Segmentation (BraTS) 2020 data \cite{ref_brats1,ref_brats2,ref_brats3}. We split the 369 publicly available samples into a hold-out test-set of 78 subjects, a validation-set of 42 subjects, and a training-set of 249 subjects. All contain both high-grade and low-grade glioma cases.

\subsection{Experiments and Uncertainty Measures}

We have trained a 2.5-dimensional U-net to segment each tumor class (i.e., whole tumor, tumor core, and enhancing tumor) using: 6-block encoder-decoder pathway, 8 initial filters (doubling per level), Adamax optimiser ($\beta_1,\beta_2 = 0.9,0.999$), learning rate schedule starting 0.001 reducing by 0.25 factor after patience/cooldown of 3/2 epochs, 50 training epochs, cross-entropy/soft-Dice loss weighted 0.3/0.7, and 2.5D input stacks with 2 adjacent axial slices each side per reference slice.

For each TTD and TTA case (as in sections \ref{subsec:TTD} and \ref{subsec:TTA}), each test-set sample is evaluated to produce voxelwise probabilities (of belonging to each class). Each sample is processed 50 times with random dropout or data augmentation applied on each forward pass, creating separate distributions, in each voxel, for each uncertainty case. From each distribution, the mean, variance and entropy are calculated and stored as voxelwise maps for each uncertainty case.

To assess how consistent and repeatable these uncertainty values are, we calculated the median and interquartile range, in each voxel separately, as stable measures should have a small range of values. These robust measures would not be substantially affected by one or two unusual/outlier cases.

In addition, to assess whether there is a range of different spatial patterns generated by different settings we measure the spatial correlation, being independent of the global magnitude. This was done for each combination of TTD and TTA cases by calculating the correlation of uncertainty spatial maps, using a mask, within each subject, since the tumor shape and location varies with each subject and so we cannot mix spatial maps across subjects. The spatial mask consisted of all voxels with a non-zero median entropy value (across all cases), thereby excluding a large number of background voxels that otherwise would inflate correlation values. These values form a correlation matrix for each subject. Additionally, we also calculated the mean of these matrices across subjects, allowing us to assess: (i) if the relationships between spatial patterns from different uncertainty sources are consistent or not, and (ii) if more than a single map is needed to capture their diversity or not.

Finally, we assess the magnitude of the uncertainty values, independent of spatial patterns, by calculating the mean of non-zero uncertainty values for each case and subject, to capture the average uncertainty level. We also trained an auxiliary network to predict the error likelihood of the primary segmentation network (``predicted error network") for comparative purposes and as an analogue to the model calibration methods highlighted in section~\ref{sec:related_work}; here, we use the primary network voxel-wise classification confidence levels to define a loss that is minimised when the predicted error network outputs equate to the distance between ground-truth labels and primary network confidence levels. 

\section{Results}

For brevity, we report results only for tumor core tissue using voxelwise entropy values as our measure of uncertainty. These results generalise to using the variance as the uncertainty measure, as well as to other tumor tissue categories. The U-net performance (Dice=0.74) is not the focus but is comparable with the mean of BraTS 2020 challenge submissions.

\subsection{Voxelwise Uncertainty with TTD/TTA Parameter Variations}

\begin{figure}[tb]
\centerline{\includegraphics[scale=0.62]{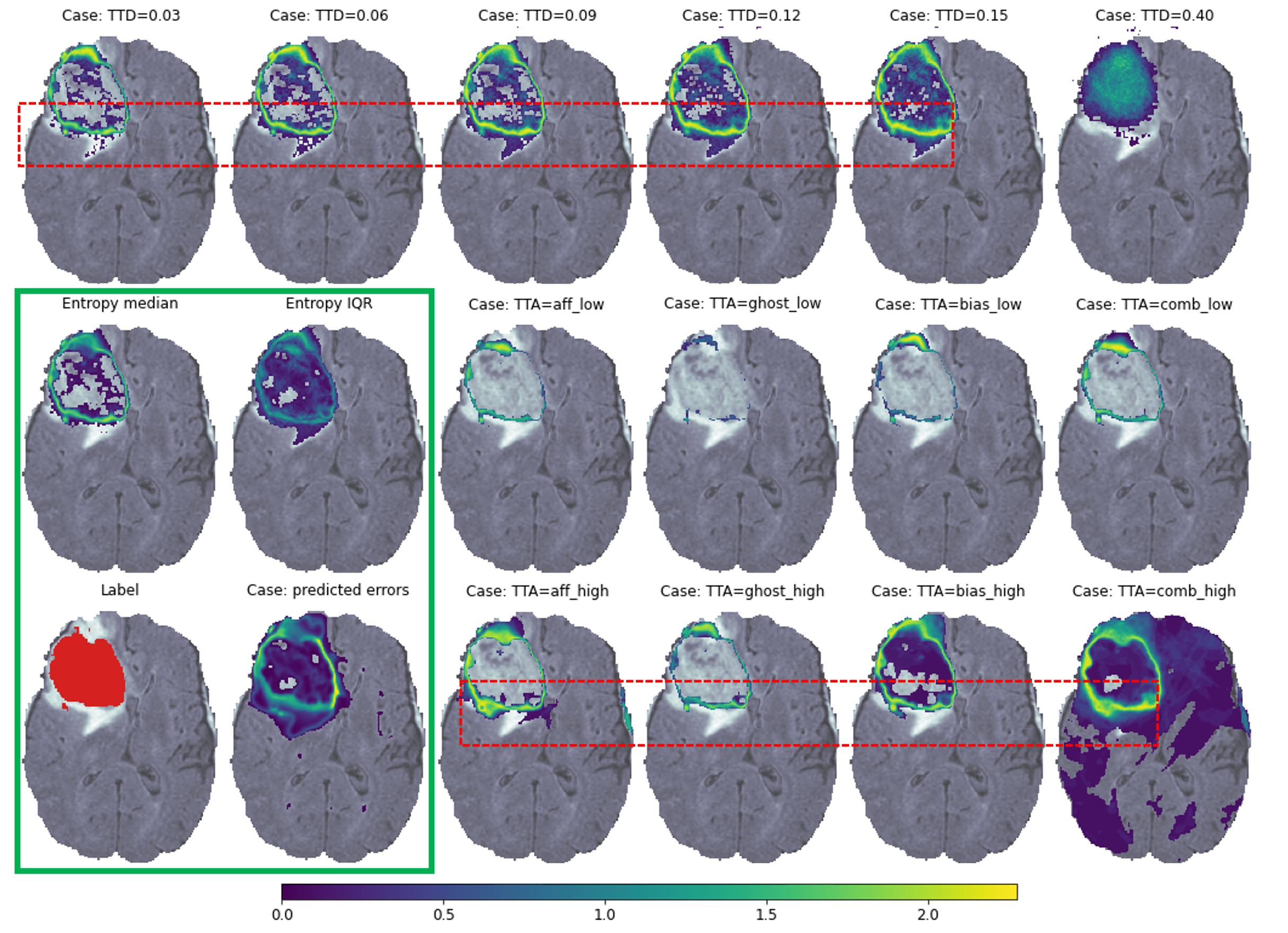}}
\caption{Subject 197, axial slice 81/155: entropy (uncertainty) maps for different TTD/TTA cases (see sections \ref{subsec:TTD} and \ref{subsec:TTA}); along with the median and interquartile range across all uncertainty cases, a predicted error map for the underlying network and the ground-truth label in the green highlight box. The maximum predicted error value (bright yellow) is 0.248 for this axial slice. The dashed red boxes highlight one example of the diversity of uncertainties derived from various sources---here, the affine, bias-field and combined aleatoric cases express considerably greater uncertainty in the small discontiguous tumor region (shown in the lower left part of the main label region for this axial slice) as compared to the TTD cases.} \label{fig1}
\end{figure}

Figure~\ref{fig1} provides examples of individual entropy maps for all uncertainty cases, as well as the median and interquartile range, in a randomly selected test subject (ID: 197). Across the TTD/TTA parameter settings, it is clear that median entropy levels are reasonably consistent, as we would want them to be. Perhaps more interesting is the interquartile range, where it is clear that even with moderate parameter settings, there is a substantial spread of values in many locations. Purely for comparative purposes, a predicted error likelihood map is also produced using the predicted error network and is displayed in figure~\ref{fig1} showing a similar, but also slightly different, spatial map. 

\subsection{Spatial Correlations}

\begin{figure}[t]
\centerline{\includegraphics[scale=0.35]{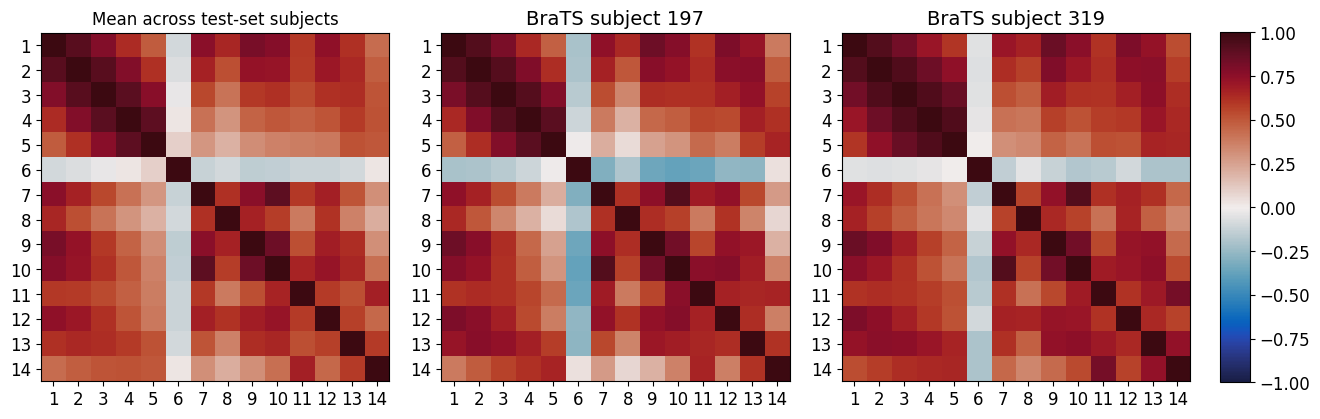}}
\caption{Correlation matrices (of spatial correlations) for the entropy maps generated from the different TTD and TTA uncertainty cases. The first is the mean across all test subjects, while the others are individual subjects (IDs: 197 and 319).
Rows and columns represent the different uncertainty cases in the following order: Cases 1--6 epistemic TTD probability = [0.03, 0.06, 0.09, 0.12, 0.15, 0.40] respectively; Cases 7--14 aleatoric TTA = [7: affine low, 8: ghosting low, 9: bias-field low, 10: combined (affine, ghosting, bias-field) low, 11: affine high, 12: ghosting high, 13: bias-field high, 14: combined (affine, ghosting, bias-field) high]. Section \ref{subsec:TTA} provides TTA details.} \label{fig2}
\end{figure}

Matrices of spatial correlations between uncertainty cases are shown in figure~\ref{fig2}. These quantify the similarity of spatial patterns, independent of the location and size of the tumors. Average correlation matrices, across subjects, show if the high and low uncertainty regions are consistently placed.

It can be seen from these results that there are clear similarities among the first 5 TTD cases, although the correlation values reduce as probability settings diverge. A striking result is the negative and near-zero correlation exhibited between case 6 (TTD probability of 0.40) and each of the other TTD and TTA cases. Case 6 provides significant disruption to the network such that non-zero uncertainty values are spread across the whole image (including non-brain regions). Correlations between TTA cases are more varied with a less discernible pattern. This is not unexpected given the variety of data manipulations that are possible using various techniques. Finally, correlations between TTD and TTA cases show a discernible, repeatable pattern, with correlation dropping as the TTD probability parameter is increased; signifying that while low probability TTD cases have similar patterns to aleatoric cases, as this probability increases the spatial pattern of high and low uncertainty regions for these estimates of epistemic and aleatoric uncertainty become increasingly different. 

\subsection{Mean Uncertainty Levels}

\begin{figure}[t]
\centerline{\includegraphics[scale=0.31]{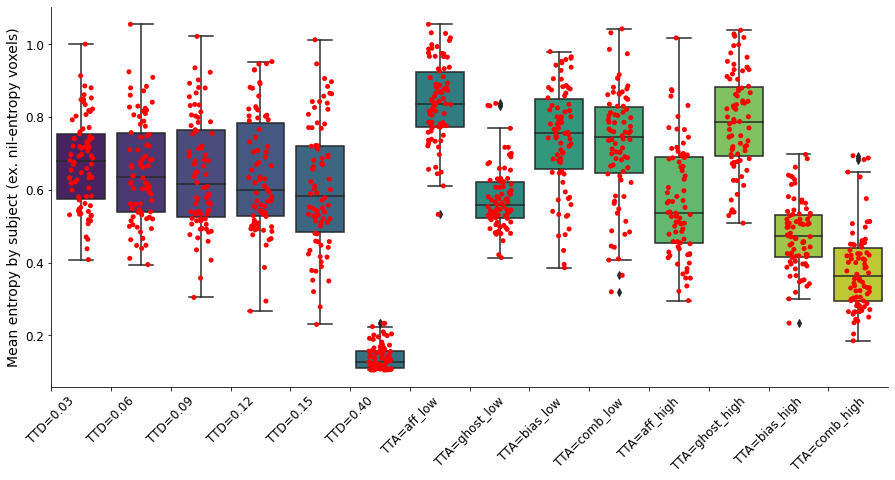}}
\caption{Boxplots of the mean of non-zero voxelwise entropy values, for each subject, shown for epistemic (TTD) and aleatoric (TTA) uncertainty cases. For TTD case 6, the large network disruption led to many voxels with low, but non-zero, entropy values.} \label{fig3}
\end{figure}

Since the spatial correlations are invariant to the global magnitude of the uncertainties, we have also measured the mean of non-zero voxel entropy values for each subject, for each uncertainty case, to quantify global magnitude. These results can be seen in figure~\ref{fig3}. Once again, the extreme TTD case (i.e., case 6) is a conspicuous outlier, with much lower values, caused by many small values that are spread across the image. This provides further confirmation that TTD settings with higher probability values are unlikely to be of use. The remaining 5 TTD cases behave similarly, with wider distribution of mean levels for higher dropout probabilities. For TTA cases, the high bias-field and high-level combined TTA case show lower distributions of means than the other TTA cases. 

\section{Discussion and Conclusion}

It is clear from these results that uncertainty distributions for epistemic (TTD) and aleatoric (TTA) sources are distinctly different from each other and that the parameter types and values within them also have a substantial impact on the uncertainty distributions. This indicates that more than one category of uncertainty should be modeled and estimated for a variety of use cases, such as to (i) aid interpretability, (ii) capture a diversity of uncertainty sources, (iii) indicate reliability, or (iv) incorporate into uncertainty-aware networks. Although reported results only used entropy, these conclusions also hold for variance and other tissue classes. We chose to show entropy values as they are better for distributions that are not unimodal, which were likely to be produced. In this early work we chose not to include all uncertainty sources (e.g., inter-rater variability) or compare to further techniques such as ensembles or Bayesian-based network architectures, but rather demonstrate the diversity of uncertainties generated by two common methods - especially accounting for MR artefacts with TTA.

In some instances, the interquartile range approached the full range of possible entropy values. This shows that the parameters used for TTD or TTA uncertainty estimation are important. For more extreme dropout rates the TTD uncertainty estimates are not useful for any purpose and so researchers should err on the side of using reduced dropout probability settings. Furthermore, the calculated uncertainties may gradually change from being useful to non-informative, based on the gradual changes in correlations found at lower dropout values, although further testing would be required to establish this generally.

Aleatoric uncertainties spanned a range of distinctly different maps, as shown by the correlation matrices, indicating that a number of different data augmentation types need to be included to capture aleatoric uncertainty. The augmentation parameters also clearly have an effect, and when pushed to higher values can sometimes result in widespread low-grade uncertainties across images. This could be used to establish a practical upper-bound for these parameter settings.

Careful consideration of different parameter settings is needed when using TTD and TTA estimation methods, and researchers should consider that there is no single uncertainty map, but rather a set of different distributions governed by the TTD and TTA hyper-parameter values and types. In order to have reliable, repeatable, interpretable measures of uncertainty it is important to specify the types of augmentation and all hyper-parameters. When using uncertainties with uncertainty-aware networks, we expect incorporating maps from multiple categories to be beneficial. Finally, while a predicted error map may share similarities with some uncertainty cases, they measure at most one aspect of uncertainty and possibly something different again, and are thus not a replacement for a \textit{set} of maps from different uncertainty categories. Researchers should carefully consider parameter settings for each method, as figure~\ref{fig3} demonstrates how larger parameter settings can produce unhelpful, unrepresentative uncertainties, indicating that upper-bounds should be established in each case (e.g., TTA cases: bias-field and combined).

In this study one major limitation is the size and nature of the dataset. BraTS is a well studied and characterised dataset, which contains several different segmentation tasks, and thus allowed us to verify that the findings generalised to different tissue types and also different uncertainty measures. However, to establish these results more generally, further studies are required.

In conclusion, we have shown that there are multiple sources of uncertainties that generate distinctly different uncertainty maps (see figure~\ref{fig2}), and that these should be incorporated into all uncertainty work. The epistemic and aleatoric uncertainty estimates, as obtained through TTD and TTA respectively, are sensitive to their hyper-parameter settings and to obtain repeatable maps it is important to specify the hyper-parameters involved. These settings need to be thought through carefully when building and running networks that produce uncertainty estimates. Furthermore, we found that there is a richness in the set of uncertainty estimations that is not captured by a single distribution, and this should be considered when feeding maps into uncertainty-aware networks, or when using them to aid interpretability or indicate reliability. Therefore, although a chosen distribution or summary may capture uncertainties to some degree, it is unlikely to provide comprehensive estimates across all regions where uncertainties could manifest; which could be vital to ensuring clinical efficacy in challenging situations.

%
%
%

\begin{thebibliography}{8}

\bibitem{ref_dinsdale}
Dinsdale, N. et. al.: Challenges for machine learning in clinical translation of big data imaging studies. arXiv:2107.05630 (2021)


\bibitem{ref_gal}
Gal, Y., Ghahramani, Z.: Dropout as a Bayesian approximation: Representing model uncertainty in deep learning. In: Proceedings of the 33\textsuperscript{rd} International Conference of Machine Learning -- vol. 48, ICML'16, pp. 1050--1059. JMLR.org, (2016)

\bibitem{ref_taye}
Taye, M., Azizpour, H., Smith, K.: Bayesian uncertainty estimation for batch normalized deep networks. In.: International Conference on Machine Learning, pp. 4907--4916. PMLR (2018)

\bibitem{ref_neal}
Neal, R.M.: Bayesian learning for neural networks. Springer Science \& Business Media (2012)

\bibitem{ref_ayhan}
Ayhan, M.S., Berens, P.: Test-time augmentation for estimation of heteroscedastic aleatoric uncertainty in deep neural networks (2018)

\bibitem{ref_wang}
Wang, G. et. al.: Aleatoric uncertainty estimation with test-time augmentation for medical image segmentation with convolutional neural networks. Neurocomputing \textbf{338}, pp. 34--34 (2019) 

\bibitem{ref_ozdemir}
Ozdemir, O. et. al.: Propogating uncertainty in multi-stage Bayesian convolutional neural networks with application to pulmonary nodule detection. arXiv:1712.00497 (2017)

\bibitem{ref_herzog}
Herzog, L. et. al.: Integrating uncertainty in deep neural networks for MRI based stroke analysis. In: Medical Image Analysis \textbf{65}, DOI:10.1016/j.media.2020.101790 (2020)

\bibitem{ref_wang2}
Wang, G. et. al: Automatic brain tumor segmentation based on cascaded convolutional neural networks with uncertainty estimation. In: Frontiers in Computational Neuroscience \textbf{13}(56), DOI:10.3389/fncom.2019.00056 (2019)

\bibitem{ref_arega}
Arega, T.W. et. al.: Leveraging Uncertainty Estimates to Improve Segmentation Performance in Cardiac MR. In: MICCAI Uncertainty for Safe Utilization of Machine Learning in Medical Imaging (UNSURE), and Perinatal Imaging, Placental and Preterm Image Analysis, pp. 24--33. Springer, Cham (2021)

\bibitem{jungo_1}
Jungo, A., Reyes, M. . Assessing Reliability and Challenges of Uncertainty Estimations for Medical Image Segmentation. In: Medical Image Computing and Computer Assisted Intervention – MICCAI 2019. MICCAI 2019. Lecture Notes in Computer Science, vol 11765. Springer, Cham, DOI:10.1007/978-3-030-32245-8\_6 (2019)

\bibitem{jungo_2}
Jungo, A. et. al.: In: Analyzing the Quality and Challenges of Uncertainty Estimations for Brain Tumor Segmentation. In: Frontiers in neuroscience, 14, 282, DOI:10.3389/fnins.2020.00282 (2020)

\bibitem{kendall}
Kendall, A., and Gal, Y.: What uncertainties do we need in Bayesian deep
learning for computer vision?. In: Advances in Neural Information Processing
Systems 30 (Curran Associates, Inc.), 5574–5584 (2017)

\bibitem{mehrtash}
Mehrtash, A. et. al.: Confidence Calibration and Predictive Uncertainty Estimation for Deep Medical Image Segmentation. In: IEEE Transactions on Medical Imaging, vol. 39, no. 12, pp. 3868-3878, Dec. 2020, DOI:10.1109/TMI.2020.3006437 (2020)

\bibitem{rousseau}
Rousseau, A.-J. et. al.: Post Training Uncertainty Calibration Of Deep Networks For Medical Image Segmentation. In: IEEE 18th International Symposium on Biomedical Imaging (ISBI), 2021, pp. 1052-1056, DOI:10.1109/ISBI48211.2021.9434131 (2021)

\bibitem{ref_kingma}
Kingma, D., Welling, M.: Auto-encoding Variational Bayes. arXiv:1312.6114 (2013)

\bibitem{ref_kohl}
Kohl, S. et. al.: A Probabilistic U-Net for Segmentation of Ambiguous Images. arXiv:1806.05034 (2018) 

\bibitem{ref_baumgartner}
Baumgartner, C. et. al.: PHiSeg: Capturing Uncertainty in Medical Image Segmentation. arXiv:1906.04045 (2019)

\bibitem{ref_brats1}
Menze, A. et. al.: The Multimodal Brain Tumor Image Segmentation Benchmark (BRATS). In: IEEE Transactions on Medical Imaging \textbf{34}(10), 1993-2024, DOI:10.1109/TMI.2014.2377694 (2015)

\bibitem{ref_brats2}
Bakas, H. et. al.: Advancing the Cancer Genome Atlas glioma MRI collections with expert segmentation labels and radiomic features. In: Nature Scientific Data \textbf{4}(170117), DOI:10.1038/sdata.2017.117 (2017)

\bibitem{ref_brats3}
Bakas, S. et. al.: Identifying the Best Machine Learning Algorithms for Brain Tumor Segmentation, Progression Assessment, and Overall Survival Prediction in the BRATS Challenge. arXiv:1811.02629 (2018)

\end{thebibliography}
%

\end{document}